\documentclass[runningheads]{llncs}

\usepackage{eccv}

\usepackage{eccvabbrv}
\usepackage[table]{xcolor}
\usepackage{rotating}

\usepackage{graphicx}
\usepackage{multirow}
\usepackage{booktabs}

\usepackage[accsupp]{axessibility}  
\usepackage{fontawesome5}
\newcommand{\mailicon}{\faEnvelope}

\usepackage{hyperref}

\usepackage{orcidlink}

\begin{document}

\title{VIVAS: Vitalizing Visual Perception in VLM Pre-training via Vision-language Unified Autoregressive Supervision}

\titlerunning{VIVAS}

\vspace{-50mm}

\author{Zhehan Kan\inst{1,  3,}\textsuperscript{\textdagger} \and
Yubo Zhu\inst{2,3,}\textsuperscript{\textdagger}\and
Xinghua Jiang\inst{3,}\textsuperscript{\textdagger}\and
Zhixiang Wei\inst{3,}\textsuperscript{\textdagger}\and \\
Shifeng Liu\inst{3,}\textsuperscript{\textdagger}\and
Wei Tong\inst{2,}\textsuperscript{\mailicon} \and
Sheng Zhong\inst{2} \and
Qingmin Liao\inst{1} \and\\
Wenming Yang\inst{1,}\textsuperscript{\mailicon} \and
Xin Li\inst{3,}\textsuperscript{\mailicon} \and
Yinsong Liu\inst{3} \and
Deqiang Jiang\inst{3} \and
Xing Sun\inst{3}
}

\authorrunning{Z. Kan et al.}
\vspace{-5mm}
\institute{$^{1}$Tsinghua University \quad $^{2}$Nanjing University \quad $^{3}$Tencent Youtu Lab}

\maketitle

\begin{tikzpicture}[remember picture, overlay]
\node[
    anchor=north west,
    align=left,
    text width=\textwidth
] at ([xshift=\oddsidemargin+1in,yshift=4.3cm]current page.south west) {%

    \rule{\textwidth}{0.4pt}
    \footnotesize
    \textsuperscript{\textdagger} Equal contribution. Work done during Zhehan Kan and Yubo Zhu's internship at Tencent Youtu Lab.\\
    \textsuperscript{\mailicon} Corresponding author
};
\end{tikzpicture}

\vspace{-8mm}

\begin{abstract}
While Vision–Language Models (VLMs) demonstrate strong capabilities, they continue to suffer from a critical limitation: insufficient fine-grained visual perception, which fundamentally limits their multimodal understanding. We attribute this bottleneck to text-dominant optimization biases during pre-training, which encourage the model to overlook fine-grained visual details, thereby limiting the capability of multimodal understanding. We investigate that overcoming this bottleneck requires two key elements: (1) a unified token space paradigm that ensures stable training dynamics, and (2) a modality-aligned dense visual supervision signal enriched with both structural granularity and semantic information to capture critical visual representations. Based on these insights, we propose VIVAS, a framework built upon the unified token space paradigm, which introduces a dense-structural–semantic vision tokenizer, which expands the textual vocabulary into a unified vision–language vocabulary by incorporating a visual vocabulary. During pretraining, VIVAS performs vision–language unified autoregressive supervision over both visual details and linguistic content, thereby enhancing visual perception to improve multimodal understanding. Trained end-to-end on 12.4T tokens, VIVAS achieves state-of-the-art performance across 7 tasks and 39 multimodal benchmarks.
 \vspace{-2mm}
\end{abstract}

\vspace{-10mm}

\begin{figure*}[htbp]
  \centering
  \includegraphics[width=0.8\linewidth]{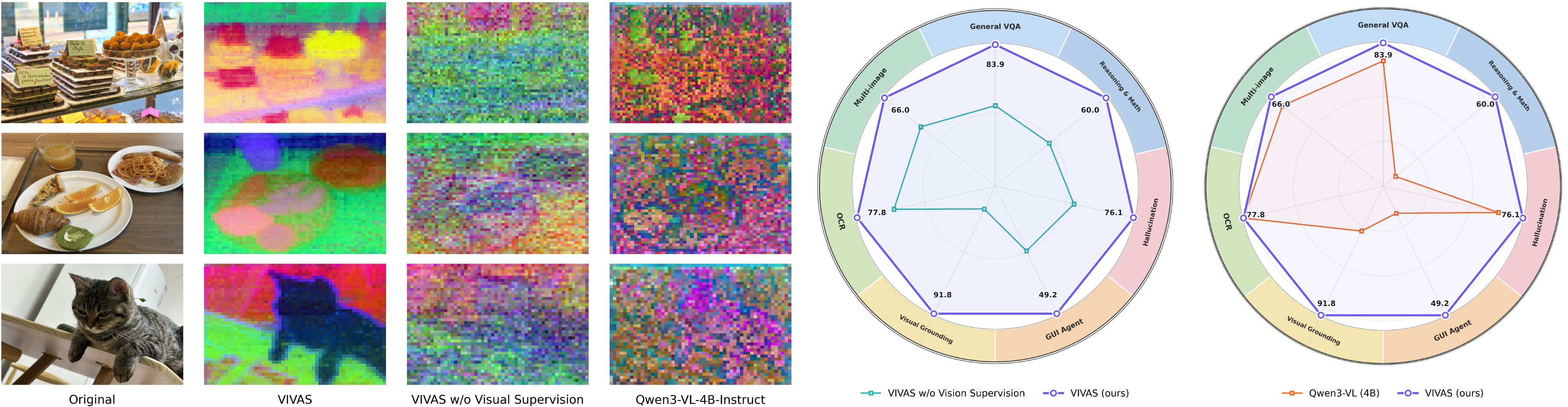}
  \caption{\textbf{Left:} Principal Component Analysis (PCA) visualizations of hidden states extracted from the output-layer vision tokens, VIVAS demonstrates significantly more fine-grained visual perception. \textbf{Right:} Comprehensive multimodal understanding evaluation of VIVAS against the SOTA Qwen3-VL (4B) and an ablated baseline without visual supervision. VIVAS yields substantial performance improvements.}
  \label{fig: teaser}
\end{figure*}

\section{Introduction}

Vision-Language Models (VLMs) have emerged as a promising paradigm for advancing toward artificial general intelligence. Conventionally, these architectures rely on a vision encoder and an alignment module to project visual representations into the linguistic space, leveraging a Large Language Model (LLM) for subsequent decoding~\cite{bai2025qwen2, chen2024expanding, liu2023visual, kan2024catchcomplementaryadaptivetokenlevel, kan2025tacothinkanswerconsistencyoptimized, zhu2025llmknowsestimatingllmperceived, kan2026rar}. Despite their widespread success, current VLMs exhibit a critical limitation: a severe deficiency in fine-grained visual perception that intrinsically bounds their multimodal understanding to a coarse level. We argue that this fundamental flaw originates from a pervasive text-dominant optimization bias inherent in standard training paradigms, where visual signals are treated merely as passive conditioning inputs. Consequently, models are implicitly incentivized to discard visual details deemed redundant for high-level text generation. This inevitably creates a visual information bottleneck, severely impeding their dense perception capabilities.

To address this, recent efforts have sought to enhance the visual perception of VLMs by injecting visual supervision during the post-training stage~\cite{li2025vista, wang2024reconstructive,yoon2025visual,li2025spatial,li2025unleashing,wang2025autoregressive}. However, at this late phase, the model's visual representations have already been deeply entrenched by text-dominant, large-scale pretraining. Moreover, since visual supervision acts merely as an auxiliary alignment constraint during post-training, its impact is inherently limited, yielding marginal improvements. Therefore, we posit that the critical key to unlocking fine-grained visual perception lies in introducing visual supervision into the pre-training stage, thereby intrinsically optimizing visual representation learning fundamentally.

\begin{figure*}[htbp]
  \centering
  \includegraphics[width=1.0\linewidth]{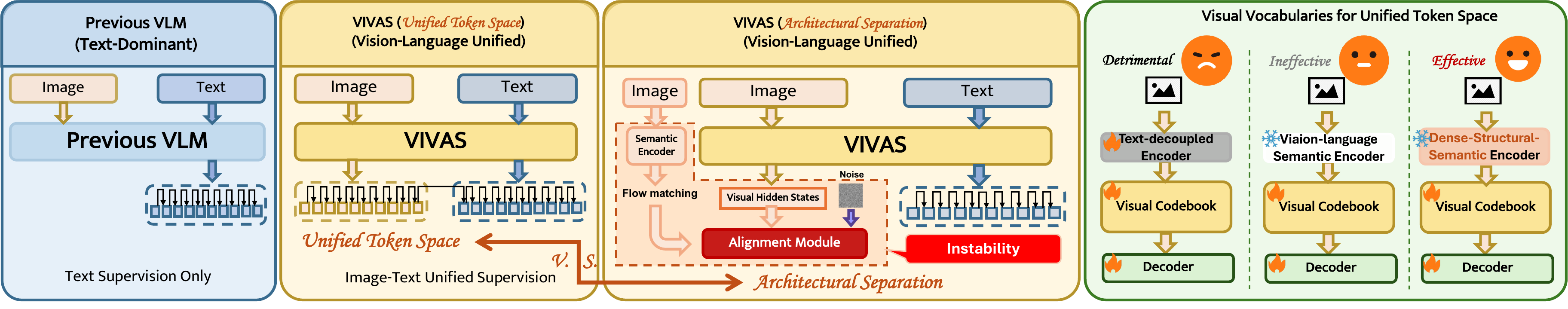}
  \caption{\textbf{Left}: Previous text-dominant VLMs lack direct visual supervision. \textbf{Middle}: To introduce visual supervision, VIVAS formulates a unified token space, fundamentally avoiding the instability inherent in architectural separation approaches. \textbf{Right}: Within this unified space, we introduce the dense-structural-semantic visual supervision signal, starkly outperforming detrimental pixel-level or ineffective semantic-level signals.}
  \label{fig: intro}
\end{figure*}

In this paper, as illustrated in Figure~\ref{fig: intro}, we first investigate the architectural paradigm requirements for effective visual supervision. We observe that the architectural separation paradigm, which introduces alignment modules for visual reconstruction during pretraining, leads to training instability and convergence difficulties. In contrast, the unified token space paradigm, built upon a standard architecture, expands the textual vocabulary into a unified vision–language vocabulary by incorporating a visual vocabulary. This expansion aligns visual tokens within a cohesive vision–language predictive stream, enabling stable and rapid convergence.

Furthermore, we study what characteristics of a visual vocabulary can effectively enhance multimodal understanding. We find that text-decoupled visual vocabularies exhibit a substantial modality disconnect from text, which hinders unified supervision; introducing such representations into visual supervision can even degrade multimodal understanding. In contrast, vision–language semantic visual vocabularies derived from image–text contrastive learning encode sparse semantic features that largely overlap with those produced by the VLM's vision encoder, providing limited complementary information for enhancing fine-grained visual perception and capturing dense structural granularity in visual representations.

To obtain a modality-aligned dense visual vocabulary enriched with both structural granularity and semantic information, we leverage the complementary properties of two representation learning paradigms. Specifically, DINOv3~\cite{siméoni2025dinov3}, trained via self-supervised distillation, produces dense and structured visual features, while SigLIP2~\cite{tschannen2025siglip2multilingualvisionlanguage}, trained with image–text contrastive learning, provides strong semantic representations. Based on this observation, we propose a dense-structural-semantic vision tokenizer. This tokenizer fuses the dense structural features from DINOv3 with the semantic features from SigLIP2 through a cross-attention mechanism, and derives the visual vocabulary via a visual reconstruction training objective.

Built upon this visual vocabulary and the unified token space paradigm, we introduce \textbf{VIVAS}, which incorporates vision–language unified autoregressive supervision over both visual details and linguistic content during pretraining, thereby enhancing visual perception to improve multimodal understanding. We conduct end-to-end training on 12.4T tokens of data, including pre-training, supervised fine-tuning (SFT), and reinforcement learning (RL). Extensive experiments across seven tasks and 39 benchmarks demonstrate that our approach significantly improves multimodal understanding by strengthening visual perception, while incurring only 3.8\% additional inference latency.

In summary, our contributions are as follows:
\begin{itemize}
    \item We identify a key limitation of existing VLMs: a \textbf{text-dominant optimization bias} that leads to the loss of fine-grained visual details and creates a visual information bottleneck.

    \item We study architectural paradigms for visual supervision in pretraining and show that a \textbf{unified token space paradigm} enables stable training by integrating visual tokens into a shared vision–language predictive stream.

    \item We propose a \textbf{dense-structural-semantic vision tokenizer} that fuses DINOv3 structural features and SigLIP2 semantic features via cross-attention to construct a modality-aligned visual vocabulary.

    \item We introduce \textbf{VIVAS}, a vision–language pre-training framework with unified autoregressive supervision, achieving significant improvements on 7 tasks and 39 benchmarks with only 3.8\% additional latency.
\end{itemize}
\section{Related Work}

\subsection{Multimodal Understanding}
Multimodal understanding refers to the capability of computational models to process, integrate, and interpret semantic information across diverse data modalities. To facilitate this paradigm, VLMs, including Seed1.5-VL, GLM-4.5, Qwen3-VL, and InternVL3.5~\cite{liu2023visualinstructiontuning, bai2025qwen3vltechnicalreport, wang2025internvl35advancingopensourcemultimodal, guo2025seed15vltechnicalreport, 5team2025glm45agenticreasoningcoding}, predominantly rely on pre-trained vision encoders (e.g., CLIP and SigLIP~\cite{zhai2023sigmoidlosslanguageimage, radford2021learningtransferablevisualmodels}) to extract continuous visual features. Optimized via contrastive learning on extensive image-text pairs, these encoders establish foundational cross-modal alignments. However, while projecting these features into the LLM's latent space enables conditional text generation, this inherently text-centric optimization severely bottlenecks the retention of dense visual details.

\subsection{Visual Supervision incorporated in VLMs}

Recent research explores the incorporation of visual supervision into VLMs to enhance visual understanding during the post-training phase. One trajectory of inquiry introduces this supervision via vision-text alignment~\cite{li2025vista}, whereas an alternative paradigm focuses on incorporating visual supervision by predicting or reconstructing image features produced by a vision tokenizer~\cite{wang2024reconstructive, yoon2025visual, li2025spatial, li2025unleashing, wang2025autoregressive, wang2026autoregressivesemanticvisualreconstruction}. However, the efficacy of these methodologies is inherently constrained. Given that the latent representations are predominantly crystallized by text-centric large-scale pretraining prior to this stage, visual signals function merely as auxiliary alignment, yielding marginal improvements. Conversely, the initial pretraining phase constitutes the critical window for fundamentally optimizing the representational capacity of VLMs. To address these limitations, we propose VIVAS, a framework that leverages dense semantic visual signals to formulate a vision-language unified autoregressive supervision, thereby significantly enhancing fine-grained visual perception.

\section{Methodology}

\subsection{Preliminary}
Given a standard VLM parameterized by $\theta$, visual inputs $v$ and text queries $x$ are mapped into a shared embedding space via a vision encoder and an alignment module. The model autoregressively generates a response.
\begin{equation}
    y_t \sim p_\theta(y_t \mid v, x, y_{<t}) \propto \exp \big[ \textit{logit}_\theta(y_t \mid v, x, y_{<t}) \big].
\end{equation}
In this paradigm, optimizing solely via text-generation objectives treats visual signals as passive conditions. Consequently, models tend to discard fine-grained visual details unnecessary for coarse-grained text generation, creating an information bottleneck that impedes dense visual perception capabilities. To overcome this limitation, Section~\ref{sec:motivation} establishes the requisite visual supervision for VLM pre-training. Subsequently, Section~\ref{sec:methodology} introduces how to obtain such visual supervision, and Section~\ref{sec:training_curriculum} details our proposed unified vision-language autoregressive training pipeline. Crucially, our pre-training entails the end-to-end optimization of the vision encoder, projector, and LLM decoder.

\subsection{What Visual Supervision Are Needed for VLM Pre-training?}
\label{sec:motivation}

To jointly ensure stable optimization while capturing essential visual representations, this section investigates the mechanisms required for visual supervision during pretraining from two primary perspectives: \textit{the visual supervision architectural paradigm} and \textit{the characteristics of the visual supervision signal}.

\begin{figure*}[htbp]
  \centering
  \includegraphics[width=1.0\linewidth]{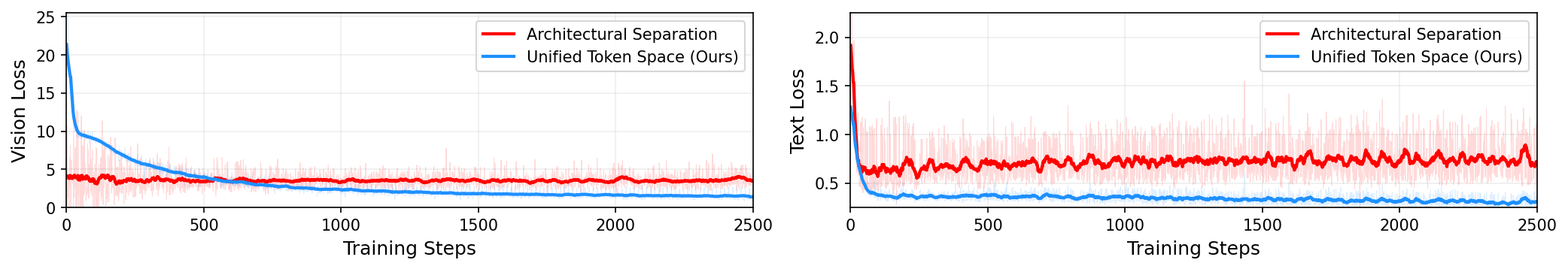}
  \caption{Comparison of training loss between architectural separation and unified token space. Vision and text losses correspond to the supervised training objectives for the visual and textual modalities, respectively.}
  \label{fig: loss_compare}
  \vspace{-5mm}
\end{figure*}

\subsubsection{Visual Supervision Architectural Paradigm.} In visual supervision for VLM pre-training, two different architectural paradigm formulations can be adopted: architectural separation and unified token space.

\textit{Architectural Separation:} Under the architectural separation paradigm, visual supervision is introduced through an auxiliary alignment module attached to the backbone VLM. Specifically, the output-layer hidden states produced by the model are fed into this module, which maps them to predefined visual targets using a feature-space regression loss. The alignment module is optimized jointly with the backbone during pre-training.

\textit{Unified Token Space:} In the unified token space paradigm, the textual vocabulary is expanded into a unified vision–language vocabulary by incorporating a visual vocabulary. The VLM is then trained autoregressively over this unified vocabulary, where visual supervision is formulated as next-token prediction under a standard cross-entropy objective.

As illustrated in the Fig.~\ref{fig: loss_compare}, evaluating the training loss trajectories highlights a fundamental divergence between architectural separation and unified token space paradigms. In the architectural separation paradigms, an external Diffusion Transformer (DiT)~\cite{peebles2023scalablediffusionmodelstransformers} is appended to the hidden states of the VLM's output-layer. These hidden states function as conditions to guide the denoising process of corrupted images. Empirical observations indicate that the architectural separation paradigm suffers from severe training instability and convergence difficulties. This primarily stems from the early stage of pretraining, when the model’s visual capability has not yet been established; introducing additional training modules at this stage to provide visual supervision disrupts optimization and exacerbates training instability. In contrast, optimization within a unified token space yields stable and rapid convergence. Consequently, these findings establish the unified token space paradigm as a crucial prerequisite for achieving stable training convergence in effective visual supervision.

\newcommand{\first}[1]{\textbf{#1}}
\newcommand{\ro}[1]{\rotatebox{90}{#1}} 

\begin{table}[htbp]
\centering
\small
\caption{Performance comparison of various visual supervision strategies. All models are pre-trained on a 0.5T-token corpus and evaluated under a 5-shot setting across diverse multimodal benchmarks.}
\renewcommand{\arraystretch}{1.1}
\setlength{\tabcolsep}{6pt}
\resizebox{\textwidth}{!}{
\begin{tabular}{l | cccccccccccc}

\toprule

\multirow{1}{*}[-0.1em]{\textbf{Methods}} & 
\ro{VisuLogic} & \ro{$MMBench_{EN}$~} & \ro{RealWorldQA} & \ro{MME} & \ro{$CRPE_{EXIST}$} & \ro{$CRPE_{RELATION}$~} & \ro{$CVBench_{2D}$} & \ro{HallusionBench} & \ro{TextVQA} & \ro{DocVQA} & \ro{RefCOCO$_{\text{val}}$} & \ro{RefCOCOg$_{\text{val}}$} \\

\midrule

w/o Visual Supervision 
& 22.2 & 53.0 & 43.0 & 48.2 & 65.7 & 57.0 & 51.0 & 55.3 & 52.7 & 60.2 & 62.1 & 59.4 \\

Text-decoupled vocabularies
& 21.6 & 51.2 & 41.8 & 46.9 & 63.5 & 55.1 & 49.3 & 53.9 & 51.4 & 58.6 & 60.7 & 57.8 \\

Vision-language semantic vocabularies
& 23.6 & 53.6 & 43.2 & 47.7 & 66.2 & 57.6 & 51.4 & 54.8 & 53.1 & 59.3 & 63.0 & 60.7 \\

Ours 
& \first{24.4} & \first{54.9} & \first{46.5} & \first{49.8} & \first{72.2} & \first{59.0} & \first{56.3} & \first{56.5} & \first{54.2} & \first{63.4} & \first{65.2} & \first{63.1} \\

\bottomrule
\end{tabular}
}
\label{tab:few_shot_eval}
\vspace{-5mm}
\end{table}

\subsubsection{Characteristics of the Visual Supervision Signal.}
\label{sec:characteristics}

Under the unified token space paradigm, we categorize visual supervision signals into two groups: \textit{text-decoupled vocabularies} and \textit{vision-language semantic vocabularies}.

\textit{Text-decoupled vocabularies.}
Text-decoupled signals are learned strictly through visual supervision, without explicit text alignment. These representations typically stem from pixel-level tokenizers (e.g., VQ-VAE) or self-supervised encoders (e.g., DINOv3). Despite varied training paradigms, both approaches capture purely visual features. In this study, we extract these representations using a standard VQ-VAE vision tokenizer optimized for image reconstruction.

\textit{Vision-language semantic vocabularies.} In contrast, semantic signals derived from image–text contrastive learning encode high-level abstractions aligned with language semantics. To obtain such representations, we adapt the VQ-VAE framework by leveraging a frozen, pretrained semantic vision encoder (SigLIP2) to produce semantic features while retaining the reconstruction objective.

Standard VLM benchmarks predominantly employ zero-shot evaluation settings. However, pre-training checkpoints typically lack instruction alignment, resulting in formatting mismatches that obscure the intrinsic capabilities of the model. To decouple representation quality from instruction-following proficiency, we adopt a few-shot evaluation protocol aligned with recent assessment methodologies for foundational language models~\cite{deepseekai2025deepseekv3technicalreport}. Specifically, we leverage in-context learning to standardize output formats, thereby facilitating a rigorous assessment of the base model independent of explicit instruction tuning.

Utilizing this protocol, we investigate the impact of different visual supervision signals during VLM pre-training. As summarized in Table~\ref{tab:few_shot_eval}, incorporating either text-decoupled or vision–language semantic supervision results in suboptimal performance compared to a purely text-supervised baseline. Specifically, models trained with text-decoupled supervision exhibit a noticeable degradation in multimodal understanding. In contrast, semantic supervision derived from image–text contrastive models provides only marginal improvements and occasionally leads to slight performance drops across standard benchmarks.
These observations reveal two fundamental representational limitations. First, text-decoupled supervision, learned purely from visual signals without alignment to the language modality, introduces a substantial semantic gap that hinders alignment with high-level textual concepts. This misalignment leads to a pronounced modality disconnect. Second, vision–language semantic supervision derived from contrastive vision–language encoders generates representations that are highly correlated with those already captured by the VLM's vision encoder. Consequently, such supervision provides limited complementary information and fails to introduce the structural granularity required for improving fine-grained visual perception.

\subsection{Dense-Structural-Semantic Visual Supervision}
\label{sec:methodology}

Motivated by the limitations discussed in Section~\ref{sec:characteristics}, we aim to design a visual supervision mechanism that preserves training stability while providing a dense supervision signal enriched with both structural granularity and semantic information during pretraining. To achieve this, we propose a dense-structural-semantic vision tokenizer. This vision tokenizer initially synthesizes dense, modality-aligned representations enriched with explicit spatial anchoring, subsequently discretizing them via a compact codebook to derive the requisite visual supervisory signals.

DINOv3, optimized via self-distillation, excels in delineating local geometry, high-frequency structures, and precise object-part boundaries; nevertheless, it inherently lacks modality-aligned semantics. Conversely, SigLIP2 yields robust category-level semantics through rigorous vision-language alignment, yet its spatial representations remain intrinsically coarse. This spatial deficiency frequently precipitates blurred boundaries and semantic bleeding across adjacent regions. To synthesize visual representations that are concurrently structurally dense and semantically aligned, we leverage DINOv3 features as structural queries ($\mathbf{Q}$) to establish rigorous spatial templates, while treating SigLIP2 features as the semantic key-value repository ($\mathbf{K}, \mathbf{V}$) for the retrieval of language-grounded attributes.

\begin{figure*}[htbp]
  \centering
  \includegraphics[width=1.0\linewidth]{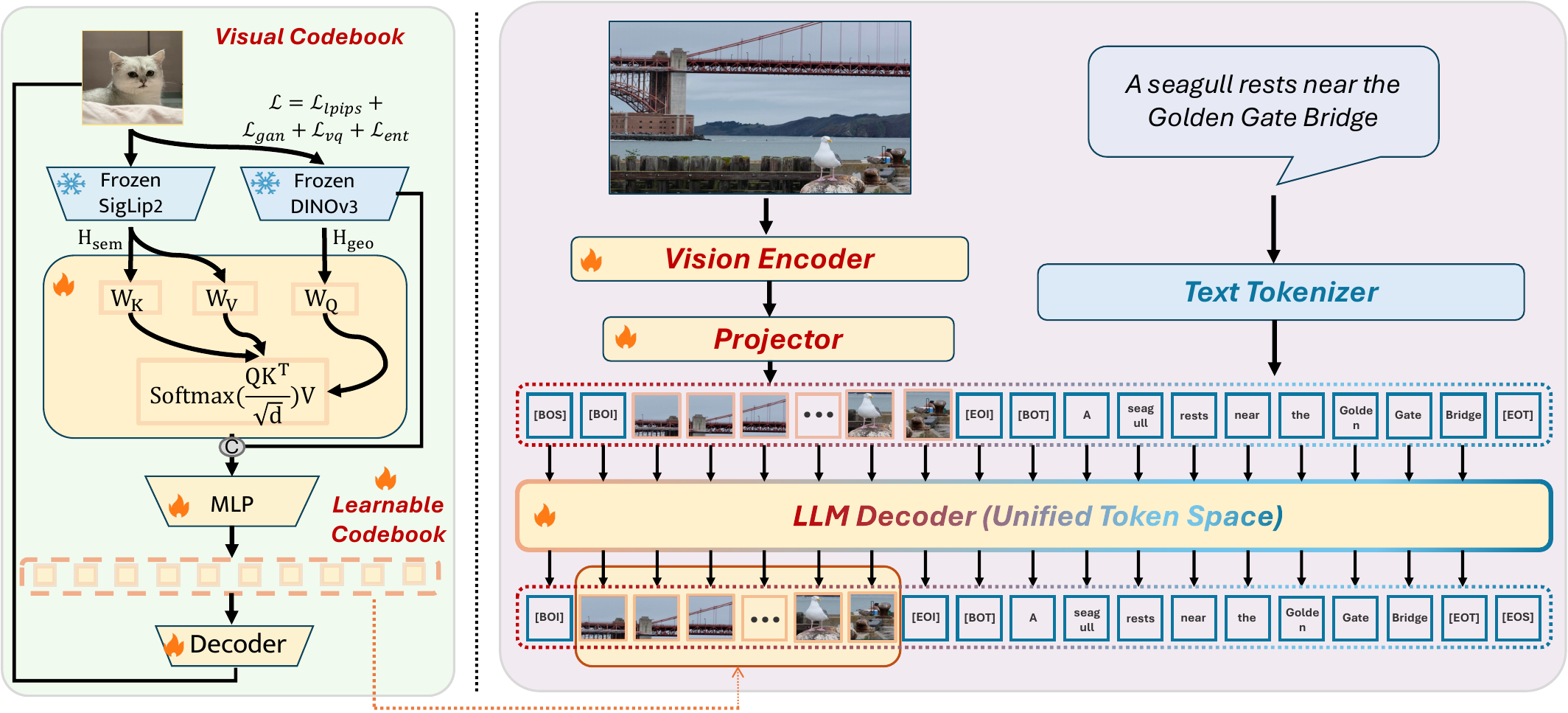}
  \caption{\textbf{Overview of VIVAS.} \textbf{Left}: The dense structural–semantic vision tokenizer constructs a visual codebook by fusing semantic and geometric features through cross-attention. \textbf{Right}: During end-to-end pretraining, VIVAS performs unified autoregressive supervision over both the textual vocabulary and the vision vocabulary.}
  \label{fig: method}
\end{figure*}

Given an input image, it is processed through dual frozen expert encoders to extract patch-level feature embeddings. Specifically, let $\mathbf{H}_{\text{geo}}\in\mathbb{R}^{N\times d_{\text{d}}}$ and $\mathbf{H}_{\text{sem}}\in\mathbb{R}^{N\times d_{\text{s}}}$ denote the sequences of DINOv3 tokens and SigLIP-2 tokens, capturing geometry-aware and language-aligned semantic features, respectively. Here, $N$ represents the number of spatial patches, while $d_{\text{d}}$ and $d_{\text{s}}$ correspond to their feature dimensions. To integrate these representations, a cross-attention fusion mechanism is employed to probe semantic features under structural constraints. By projecting the feature maps into a shared latent space, the Query ($\mathbf{Q}$), Key ($\mathbf{K}$), and Value ($\mathbf{V}$) matrices are defined as:
\begin{equation}
    \mathbf{Q} = \mathbf{H}_{\text{geo}} \mathbf{W}_Q,\quad
    \mathbf{K} = \mathbf{H}_{\text{sem}} \mathbf{W}_K,\quad
    \mathbf{V} = \mathbf{H}_{\text{sem}} \mathbf{W}_V,
\end{equation}
whereafter the synergistic representation $\mathbf{Z}_{\text{syn}}$ is computed via cross-attention:
\begin{equation}
    \mathbf{Z}_{\text{syn}} =
    \text{Softmax}\!\left(\frac{\mathbf{Q}\mathbf{K}^{\top}}{\sqrt{d_k}}\right)\mathbf{V}.
\end{equation}
Prior to quantization, $\mathbf{Z}_{\text{syn}}$ is concatenated with the structural features $\mathbf{H}_{\text{geo}}$ along the channel dimension. The composite representation is projected via a multi-layer perceptron (MLP) and discretized using Index Backpropagation Quantization (IBQ)~\cite{shi2025scalableimagetokenizationindex}. This operation maps continuous vectors to the nearest prototype within a learnable codebook $\mathcal{C}=\{c_k\}_{k=1}^K$, characterized by a vocabulary size of $K=150{,}000$ and an embedding dimension of $D=768$.

The tokenizer is optimized end-to-end to reconstruct the input images from the discrete codes. A compound objective function $\mathcal{L}_{\text{tok}}$ is formulated to balance perceptual fidelity with codebook utilization:
\begin{equation}
    \mathcal{L}_{\text{tok}} = \underbrace{\lambda_{p}\mathcal{L}_{\text{lpips}} + \lambda_{g}\mathcal{L}_{\text{gan}}}_{\text{Perceptual \& Adversarial Fidelity}} + \underbrace{\mathcal{L}_{\text{vq}} + \lambda_{e}\mathcal{L}_{\text{ent}}}_{\text{Codebook Optimization}}
\end{equation}
where $\mathcal{L}_{\text{lpips}}$ enforces textural realism through perceptual similarity, and $\mathcal{L}_{\text{gan}}$ represents the adversarial discriminator loss. To preclude codebook collapse, a vector quantization loss $\mathcal{L}_{\text{vq}}$ is integrated alongside an entropy regularization term $\mathcal{L}_{\text{ent}}$, with loss weights empirically set to $\lambda_p=1$, $\lambda_g=1$, and $\lambda_e=0.1$. A critical design choice in this optimization strategy is the deliberate exclusion of the standard pixel-wise $\ell_1$ reconstruction loss. Minimization of the $\ell_1$ norm inherently induces a ``texture bias'', promoting the memorization of high-frequency noise while bypassing high-level semantic abstraction. By relying exclusively on perceptual and adversarial constraints, the codebook is compelled to encode structural semantics rather than mere pixel statistics. Empirically, this configuration yields a codebook utilization rate of $97.7\%$ on the composite dataset. Consequently, the learned indices form a robust visual vocabulary $\mathcal{V}_{\text{img}}$ to provide visual supervision.

\subsection{Vision-Language Unified Autoregressive Pre-training}
\label{sec:training_curriculum}

\subsubsection{Unified Training Objective.}
We formulate multimodal learning by interleaving discrete visual tokens $V = \{v_1, \dots, v_M\}$ and textual tokens $T = \{t_1, \dots, t_N\}$ into a unified sequence $\mathcal{S}$. Under an autoregressive framework, the model is optimized to maximize the likelihood of each token $s_i$ conditioned on its preceding context $s_{<i}$, regardless of modality. The overall objective is defined as:
\begin{equation}
    \mathcal{L}_{\text{total}} = \mathcal{L}_{\text{text}} + \lambda \mathcal{L}_{\text{vision}},
\end{equation}
where $\lambda$ balances the relative contribution of visual token prediction, which we empirically set to $0.5$. Both components are optimized using token-level cross-entropy:
\begin{equation}
    \mathcal{L}_{\text{text}} = - \sum_{i \in \mathcal{I}_t} \log P(t_i \mid s_{<i}; \theta), \quad 
    \mathcal{L}_{\text{vision}} = - \sum_{j \in \mathcal{I}_v} \log P(v_j \mid s_{<j}; \theta),
\end{equation}
where $\mathcal{I}_t$ and $\mathcal{I}_v$ denote the positional indices of textual and visual tokens in $\mathcal{S}$, respectively.

\subsubsection{Progressive Training Curriculum.}
To systematically endow the model with both robust reasoning and fine-grained visual grounding, the training process is organized into three sequential stages.

\textit{Stage I: Language Backbone Pre-training.}
The language backbone is first trained on large-scale pure text corpora to establish a rigorous linguistic foundation. This unimodal phase focuses on acquiring general reasoning abilities and domain-specific knowledge. During this stage, the model is optimized solely using the textual objective:
\begin{equation}
    \mathcal{L}_{Stage_1} = \mathcal{L}_{\text{text}}.
\end{equation}

\textit{Stage II, III: Multimodal Pre-training.}
In this stage, the entire architecture, including the vision encoder, projector, and LLM, is trained end-to-end to align visual and linguistic modalities. Crucially, we introduce visual supervision during this phase to guide the representation learning. By activating the joint objective:

\begin{equation}
    \mathcal{L}_{Stage_{2, 3}} = \mathcal{L}_{\text{text}} + \lambda \mathcal{L}_{\text{vision}}.
\end{equation}
By explicitly incorporating $\mathcal{L}_{\text{vision}}$, the model is compelled to simultaneously predict visual details and linguistic content. This strategy effectively mitigates the text-centric optimization bias inherent in traditional MLLM pre-training and preserves dense visual perception capabilities.

\textit{Stage IV: Multimodal Post-training.}
The final stage focuses on aligning the model with human preferences and specializing it for diverse multimodal tasks through Supervised Fine-Tuning (SFT) and Reinforcement Learning (RL). This phase refines the model’s instruction-following and conversational capabilities. To ensure robust instruction alignment and superior dialogue quality, the visual supervision is deactivated during this stage:
\begin{equation}
    \mathcal{L}_{Stage_4} = \mathcal{L}_{\text{text}}.
\end{equation}

\section{Experiment}

\subsection{Implementation Details}

\subsubsection{Training Recipe. } We adopt a four-stage training pipeline with progressively increasing data diversity and task complexity.

\textit{Stage I: Language Backbone Pre-training.} The language backbone is trained through two consecutive phases: (1) commonsense-oriented pre-training and (2) STEM- and coding-centric pre-training. These two phases collectively comprise approximately 10T tokens of pure textual data, covering general knowledge, mathematical reasoning, scientific domains and coding tasks. 

\textit{Stage II, III: Multimodal Pre-training.} In these stages, the entire architecture undergoes end-to-end training to develop robust visual perception and versatile task capabilities, processing approximately 2.4T tokens in total. The process begins with foundational pre-training (Stage II, 1.8T tokens) on a mixture of image-caption pairs, vision-centric tasks, and high-quality textual data to balance cross-modal comprehension with linguistic proficiency. Subsequently, the model undergoes task adaptation (Stage III, 0.6T tokens) on diverse multimodal instructions covering general VQA, OCR, STEM, GUI navigation, and fine-grained perception (e.g., detection, grounding). We also incorporate synthetic short Chain-of-Thought (CoT) data to bolster long-context understanding and logical reasoning.

\textit{Stage IV: Multimodal Post-training.}
\textit{SFT.}
We construct a high-quality multimodal instruction dataset through two strategies, comprising approximately 4B tokens: 
(1) mining instruction-following samples from the pre-training corpus using VLM-based quality filtering and taxonomy-guided balancing; 
(2) refining open-source datasets by regenerating detailed, reasoning-enhanced responses via a rewrite-and-expand pipeline.
\textit{RL.}
We further optimize the model with reinforcement learning in three aspects, comprising approximately 0.2B tokens: 
(i) perception RL for fine-grained visual tasks such as grounding and detection; 
(ii) reasoning RL on curated STEM data with verifiable rewards; 
(iii) general RL on diverse VQA tasks including OCR and chart understanding to improve generalization.

\newcommand{\second}[1]{{#1}}

\subsubsection{Model Components.} The architectural framework of VIVAS integrates a LLM backbone predicated on Youtu-LLM~\cite{lu2026youtullmunlockingnativeagentic} with a visual perception module driven by the SigLIP2-so400m-patch16-naflex encoder.

\subsubsection{Evaluation. }
To rigorously validate the efficacy of VIVAS, an extensive evaluation is conducted across a comprehensive suite of $39$ general multimodal benchmarks. This diverse evaluation protocol encompasses a broad spectrum of tasks, including visual grounding, object detection, instance counting, visual query answering, optical character recognition (OCR), and graphical user interface (GUI) interactions. For the ablation studies concerning the codebook configurations, specifically its vocabulary size, embedding dimension, and the applied loss function, all variant models are pre-trained on a corpus of 0.5T tokens and subsequently evaluated under a 5-shot setting.                

\begin{table}[htbp]
\centering
\small
\caption{Ablation study of applying visual supervision at different training stages.}
\renewcommand{\arraystretch}{1.1}
\setlength{\tabcolsep}{8pt}
\resizebox{\textwidth}{!}{
\begin{tabular}{cc | cccc}

\toprule

\textbf{Pre-train VS} & \textbf{Post-train VS} &
VisuLogic &
MMBench$_{EN}$ &
CRPE$_{\text{relation}}$ &
HallusionBench \\

\midrule
$\times$ & $\times$ & 23.1 & 83.2 & 69.5 & 57.3 \\
\rowcolor{gray!20} \checkmark & $\times$ & \first{25.7} & \first{83.9} & \first{72.2} & \first{59.1} \\
$\times$ & \checkmark & 23.3 & 82.9 & 69.9 & 57.7 \\
\checkmark & \checkmark & 23.6 & 83.3 & 70.1 & 58.1 \\
\bottomrule
\end{tabular}
}
\label{tab:vs_stage_ablation}

\end{table}

\begin{table}[htbp]
  \centering
  \small
  \renewcommand{\arraystretch}{1.05} 
  \setlength{\tabcolsep}{5pt} 
  
  \caption{Performance comparison between VIVAS and SOTA VLMs across multiple benchmarks. All scores adhere to official evaluation protocols. The results reproduced via our internal evaluation system are marked with `*'.}
  
  \resizebox{\textwidth}{!}{
  \begin{tabular}{l | c c |c c c | c c}
  \toprule
  
  \multirow{2}{*}{\textbf{Benchmarks}} & LLaVA-OV & Qwen3-VL & LLaVA-OV & InternVL-3.5 & Qwen3-VL & VIVAS w/o & \textbf{VIVAS} \\
  & 1.5 & instruct & 1.5 &  & instruct & Vision Sup. & \textbf{(Ours)} \\
  \midrule
  \textbf{Params} & 8B & 8B & 4B & 4B & 4B & 4B & \textbf{4B} \\
  
  \midrule
  \multicolumn{8}{c}{\textbf{General VQA}} \\
  \midrule
  MMBench$_{CN}$\cite{liu2024mmbenchmultimodalmodelallaround} & 81.0 & 84.7 & 76.9 & -- & \second{83.5} & 82.7 & \first{83.6} \\
  MMBench$_{EN}$\cite{liu2024mmbenchmultimodalmodelallaround} & 84.1 & 84.5 & \first{84.2} & 80.3 & 83.9 & 83.2 & 83.9 \\
  MMStar\cite{chen2024rightwayevaluatinglarge} & 67.7 & 70.9 & 64.9 & 65.0 & \second{69.8} & 70.4 & \first{71.1} \\
  MME (/2800)\cite{fu2025mmecomprehensiveevaluationbenchmark} & -- & -- & -- & 2272 & ~~\second{2309}$^*$ & 2360 & \first{2384} \\
  CVBench$_{2d}$\cite{tong2024cambrian1fullyopenvisioncentric} & -- & -- & -- & -- & ~~\second{79.1}$^*$ & 74.2 & \first{80.4} \\
  CVBench$_{3d}$\cite{tong2024cambrian1fullyopenvisioncentric} & -- & -- & -- & -- & ~~\second{92.4}$^*$ & 87.7 & \first{93.0} \\
  ScienceQA$_{\text{val}}$\cite{lu2022learnexplainmultimodalreasoning} & -- & -- & -- & -- & ~~\second{94.7}$^*$ & 95.8 & \first{97.0} \\
  SEEDBench$_{IMG}$\cite{li2023seedbenchbenchmarkingmultimodalllms} & 77.3 & -- & 76.6 & -- & ~~\first{77.0}$^*$ & 75.8 & \second{76.9} \\
  
  \midrule
  \multicolumn{8}{c}{\textbf{Multimodal Reasoning \& Math}} \\
  \midrule
  VisuLogic\cite{xu2025visulogicbenchmarkevaluatingvisual} & -- & 22.5 & -- & -- & \second{19.0} & 23.1 & \first{25.7} \\
  MathVista$_{\text{mini}}$\cite{lu2024mathvistaevaluatingmathematicalreasoning} & 69.6 & 77.2 & 67.9 & \first{77.1} & 73.7 & 75.1 & \second{76.5} \\
  MathVerse$_{\text{mini}}$\cite{zhang2024mathversedoesmultimodalllm} & -- & 62.1 & -- & 45.8 & \second{46.8} & 53.1 & \first{56.5} \\
  LogicVista\cite{xiao2024logicvistamultimodalllmlogical} & -- & 55.3 & -- & 41.8 & \first{53.2} & 50.9 & \second{52.4} \\
  VLMsAreBlind\cite{rahmanzadehgervi2025visionlanguagemodelsblind} & -- & 74.0 & -- & -- & \second{71.9} & 85.4 & \first{88.9} \\
  
  \midrule
  \multicolumn{8}{c}{\textbf{Hallucination}} \\
  \midrule
  HallusionBench\cite{guan2024hallusionbenchadvanceddiagnosticsuite} & -- & 61.1 & -- & 44.8 & \second{57.6} & 57.3 & \first{59.1} \\
  CRPE$_{\text{exist}}$\cite{wang2024allseeing_v2} & -- & -- & -- & -- & ~~\second{95.6}$^*$ & 95.2 & \first{96.9} \\
  CRPE$_{\text{relation}}$\cite{wang2024allseeing_v2} & -- & -- & -- & \first{75.0} & ~~{71.0}$^*$ & 69.5 & \second{72.2} \\
  
  \midrule
  \multicolumn{8}{c}{\textbf{GUI Agent}} \\
  \midrule
  ScreenSpot Pro\cite{li2025screenspotproguigroundingprofessional} & -- & 54.6 & -- & -- & 59.5 & 57.1 & \first{59.6} \\
  OSWorld\cite{xie2024osworldbenchmarkingmultimodalagents} & -- & 33.9 & -- & -- & \second{26.2} & 36.4 & \first{38.8} \\
  
  \midrule
  \multicolumn{8}{c}{\textbf{Visual Grounding}} \\
  \midrule
  RefCOCO$_{\text{val}}$\cite{yu2016modelingcontextreferringexpressions} & -- & -- & -- & 92.5 & 90.7 & 88.1 & \first{93.6} \\
  RefCOCO$_{\text{A}}$\cite{yu2016modelingcontextreferringexpressions} & -- & -- & -- & 94.3 & 92.2 & 91.4 & \first{95.2} \\
  RefCOCO$_{\text{B}}$\cite{yu2016modelingcontextreferringexpressions} & -- & -- & -- & 88.2 & 86.7 & 85.1 & \first{90.8} \\
  RefCOCO+$_{\text{val}}$\cite{yu2016modelingcontextreferringexpressions} & -- & -- & -- & 87.6 & 82.9 & 87.6 & \first{90.1} \\
  RefCOCO+$_{\text{A}}$\cite{yu2016modelingcontextreferringexpressions} & -- & -- & -- & 92.3 & 89.4 & 90.1 & \first{93.9} \\
  RefCOCO+$_{\text{B}}$\cite{yu2016modelingcontextreferringexpressions} & -- & -- & -- & 81.6 & 75.6 & 81.3 & \first{85.4} \\
  RefCOCOg$_{\text{val}}$\cite{yu2016modelingcontextreferringexpressions} & -- & -- & -- & 89.6 & 87.3 & 89.5 & \first{92.2} \\
  RefCOCOg$_{\text{test}}$\cite{yu2016modelingcontextreferringexpressions} & -- & -- & -- & 89.3 & 87.7 & 89.0 & \first{92.9} \\
  
  \midrule
  \multicolumn{8}{c}{\textbf{OCR-related Understanding}} \\
  \midrule
  AI2D$_{\text{test}}$\cite{kembhavi2016diagramworthdozenimages} & 84.2 & 85.7 & 83.6 & 82.6 & \second{84.1} & 84.2 & \first{85.6} \\
  InfoVQA$_{\text{val}}$\cite{mathew2021infographicvqa} & 78.4 & 83.1 & 76.1 & 78.0 & \first{80.3} & 77.8 & \second{79.1} \\
  TextVQA$_{\text{val}}$\cite{singh2019vqamodelsread} & -- & -- & -- & 77.9 & ~~\first{80.8}$^*$ & 79.4 & \second{79.6} \\
  DocVQA$_{\text{val}}$\cite{mathew2021docvqadatasetvqadocument} & 95.0 & 96.1 & 94.4 & 92.4 & \first{95.3} & 94.1 & \second{94.4} \\
  ChartQA$_{\text{test}}$\cite{masry2022chartqabenchmarkquestionanswering} & 86.5 & 89.6 & \first{87.1} & 86.0 & 84.6 & 84.7 & \second{85.3} \\
  OCRBench\cite{fu2025ocrbenchv2improvedbenchmark} & 829 & 896 & 800 & \second{822} & \first{881} & 804 & 813 \\
  SEEDBench2$_{\text{Plus}}$\cite{li2023seedbench2benchmarkingmultimodallarge} & 69.2 & -- & 68.9 & 69.4 & ~~\first{71.5}$^*$ & 69.8 & \second{71.3} \\
  CharXiv$_{\text{DQ}}$\cite{wang2024charxivchartinggapsrealistic} & 70.9 & 83.0 & 63.8 & 71.1 & \second{76.2} & 76.5 & \first{79.4} \\
  CharXiv$_{\text{RQ}}$\cite{wang2024charxivchartinggapsrealistic} & -- & 46.4 & -- & 39.6 & \second{39.7} & 41.6 & \first{43.8} \\
  
  \midrule
  \multicolumn{8}{c}{\textbf{Multi-image \& Real-world}} \\
  \midrule
  BLINK\cite{fu2024blinkmultimodallargelanguage} & -- & 69.1 & -- & 58.1 & \first{65.8} & 62.7 & \second{64.3} \\
  RealWorldQA & 68.1 & 71.5 & 67.8 & 66.3 & \second{70.9} & 72.9 & \first{74.6} \\
  MMERealWorld$_{EN}$\cite{zhang2025mmerealworldmultimodalllmchallenge} & 61.7 & -- & 61.6 & -- & ~~\first{63.0}$^*$ & 60.7 & \second{61.5} \\
  MMERealWorld$_{CN}$\cite{zhang2025mmerealworldmultimodalllmchallenge} & 56.1 & -- & 49.6 & 59.8 & ~~\second{61.3}$^*$ & 61.2 & \first{63.5} \\
  
  \bottomrule
  \end{tabular}
  } 
  
  \label{tab:all_multimodal_tasks_transposed}
  \end{table}

\subsection{Quantitative Results}

As delineated in Table~\ref{tab:all_multimodal_tasks_transposed}, a comprehensive evaluation across 39 diverse multimodal benchmarks reveals that VIVAS establishes SOTA performance on the most multimodal understanding tasks. Notably, the proposed method demonstrates pronounced improvements on tasks necessitating fine-grained visual perception, such as visual grounding. Furthermore, ablation comparisons against the baseline variant devoid of pre-training visual supervision confirm substantial performance gains attributable to the proposed mechanism. Crucially, these representational enhancements are achieved with strict computational efficiency; the requisite vocabulary expansion incurs a nominal inference time overhead of merely 3.8\%.

\begin{table}[htbp]
\centering
\small
\caption{Ablation study on the codebook size of the visual tokenizer.}
\renewcommand{\arraystretch}{1.1}
\setlength{\tabcolsep}{8pt}
\resizebox{\textwidth}{!}{
\begin{tabular}{l | cccc|c} 

\toprule

\textbf{Codebook Size} &
VisuLogic &
MMBench$_{EN}$ &
CRPE$_{\text{relation}}$ &
HallusionBench &
Utilization (\%) \\

\midrule

50k  & 23.6 & 53.5 & 57.3 & 55.6 & 98.3 \\
100k & 24.0 & 54.1 & 58.1 & 56.1 & 98.6 \\
\rowcolor{gray!20} 150k & \first{24.4} & \first{54.9} & \first{59.0} & \first{56.5} & 97.7 \\
200k & 24.2 & 54.5 & 58.7 & 56.3 & 96.6 \\

\bottomrule
\end{tabular}
}
\label{tab:codebook_ablation}
\end{table}

\begin{table}[htbp]
\centering
\small
\caption{Ablation study on the codebook embedding dimension of the visual tokenizer.}
\renewcommand{\arraystretch}{1.1}
\setlength{\tabcolsep}{8pt}
\resizebox{\textwidth}{!}{
\begin{tabular}{l | cccc|c}

\toprule

\textbf{Codebook Dim} &
VisuLogic &
MMBench$_{EN}$ &
CRPE$_{\text{relation}}$ &
HallusionBench &
Utilization (\%) \\

\midrule

64   & 22.7 & 53.1 & 57.4 & 55.5 & 95.1 \\ 
256  & 23.6 & 53.7 & 58.6 & 55.9 & 97.8 \\
\rowcolor{gray!20} 768  & \first{24.4} & \first{54.9} & \first{59.0} & \first{56.5} & 97.7 \\
1024 & 23.9 & 54.6 & 58.6 & 56.4 & 94.3 \\

\bottomrule

\end{tabular}
}
\label{tab:codebook_dim_ablation}

\end{table}

\subsection{Out-of-Distribution and Fine-Grained Generalization}

\begin{table}[t]
\centering
\small
\caption{Out-of-distribution and fine-grained perception evaluation.}
\renewcommand{\arraystretch}{1.05}
\setlength{\tabcolsep}{6pt}
\resizebox{\linewidth}{!}{
\begin{tabular}{lcccc}
\toprule
Benchmark & Qwen3-VL-4B & VIVAS w/o VS & VIVAS & Gain \\
\midrule
GEOBench-VLM\cite{danish2025geobenchvlmbenchmarkingvisionlanguagemodels} & 40.21 & 44.64 & 46.37 & +1.73 \\
VLMsAreBiased\cite{wang2026vlbiasbenchcomprehensivebenchmarkevaluating} & 21.31 & 18.33 & 22.16 & +3.83 \\
V*\cite{wu2023vguidedvisualsearch} & 80.63 & 78.53 & 81.15 & +2.62 \\
\bottomrule
\end{tabular}
}
\label{tab:ood_finegrained}
\end{table}

To further examine whether the improvements of VIVAS generalize beyond the curated multimodal pre-training distribution, we evaluate the model on an out-of-distribution domain and additional fine-grained perception benchmarks. As shown in Table~\ref{tab:ood_finegrained}, VIVAS consistently outperforms the baseline without visual supervision on GEOBench-VLM, VLMsAreBiased, and V*. In particular, the improvement on GEOBench-VLM, which contains satellite imagery that is underrepresented in standard multimodal training corpora, suggests that the proposed visual supervision does not merely fit the original data distribution. The gains on VLMsAreBiased and V* further support that the visual supervision strengthens fine-grained visual perception  capabilities.

\subsection{Ablation Studies}
As delineated in Table~\ref{tab:vs_stage_ablation}, an analysis is conducted to investigate the impact of integrating visual supervision across distinct training stages. Empirical evidence indicates that confining this supervision exclusively to the post-training phase yields only marginal improvements, whereas its integration across both pre-training and post-training phases provides merely modest gains. This is primarily because introducing visual supervision during post-training interferes with the model's instruction-following capabilities. Conversely, isolating the visual supervision entirely within the pre-training stage results in substantial performance enhancements. This observation robustly substantiates that the pre-training phase serves as the pivotal window for optimizing the foundational representational capacity of VLMs. 

\begin{table}[htbp]
\centering
\small
\caption{Ablation study on adding L1 loss during codebook training.}
\renewcommand{\arraystretch}{1.1}
\setlength{\tabcolsep}{8pt}
\resizebox{\textwidth}{!}{
\begin{tabular}{c | cccc} 

\toprule

\textbf{L1 Loss} &
VisuLogic &
MMBench$_{EN}$ &
CRPE$_{\text{relation}}$ &
HallusionBench  \\

\midrule

\rowcolor{gray!20} $\times$ & \first{24.4} & \first{54.9} & \first{59.0} & \first{56.5}  \\
\checkmark & 23.7 & 54.2 & 57.9 & 56.1  \\
\bottomrule
\end{tabular}
}
\label{tab:l1_loss_ablation}
\end{table}

\begin{table}[t]
\centering
\small
\caption{Comparison with a combined visual encoder baseline. The combined visual encoder baseline uses SigLIP2 and DINOv3 features with a learnable projector, but does not use visual-token supervision.}
\renewcommand{\arraystretch}{1.05}
\setlength{\tabcolsep}{5pt}
\resizebox{\linewidth}{!}{
\begin{tabular}{lcccccc}
\toprule
Method & VisuLogic & CVBench2D & MME & RefCOCO$_{\text{val}}$ & DocVQA & HallusionBench \\
\midrule
VIVAS w/o VS & 22.2 & 51.0 & 48.2 & 62.1 & 60.2 & 55.3 \\
w/ combined visual encoder & 22.6 & 50.7 & 48.5 & 60.9 & 60.5 & 54.6 \\
VIVAS & \textbf{24.4} & \textbf{56.3} & \textbf{49.8} & \textbf{65.2} & \textbf{63.4} & \textbf{56.5} \\
\bottomrule
\end{tabular}
}
\label{tab:combined_encoder_baseline}
\end{table}

To further examine the effect of modifying input-side visual features, we evaluate an encoder-side baseline that combines SigLIP2 with DINOv3 features through a learnable projector, while using the same 0.5T-token corpus setting and without visual-token supervision. As shown in Table~\ref{tab:combined_encoder_baseline} , this baseline fails to outperform VIVAS w/o visual supervision, remains clearly inferior to full VIVAS, and introduces additional inference cost. We attribute this to two factors: 1) under text-dominant supervision, simply enriching the input visual features is insufficient to learn fine-grained visual signals; and 2) incorporating text-decoupled DINOv3 features may dilute the text-aligned SigLIP2 representation, making vision-language alignment more difficult. In contrast, VIVAS keeps the standard SigLIP2-based continuous visual input unchanged and instead introduces explicit visual supervision during pre-training, thereby improving fine-grained perception without disrupting input-side alignment.

Furthermore, Tables~\ref{tab:codebook_ablation} and ~\ref{tab:codebook_dim_ablation} detail a hyperparameter analysis regarding the capacity, specifically the vocabulary size and embedding dimension, of the visual codebook derived from the synergistic vision tokenizer. The findings reveal that peak performance is achieved with a vocabulary size of 150,000 and an embedding dimension of 768, a configuration that concurrently sustains an exceptionally high codebook utilization rate of 97.7\%. 

In addition, while L1 loss serves as a foundational component in traditional codebook training, Table~\ref{tab:l1_loss_ablation} investigates the efficacy of its inclusion when optimizing specifically for visual understanding. The experimental results demonstrate that omitting the L1 loss yields superior performance. This is because the minimization of the L1 loss inherently induces a ``texture bias,'' compelling the codebook to allocate its representational capacity toward high-frequency noise at the expense of high-level semantic abstraction. 

\subsection{Analysis of Data Scaling}

\begin{figure*}[htbp]
  \centering  
  \begin{subfigure}[b]{0.41\linewidth}
    \centering    
    \includegraphics[width=\linewidth]{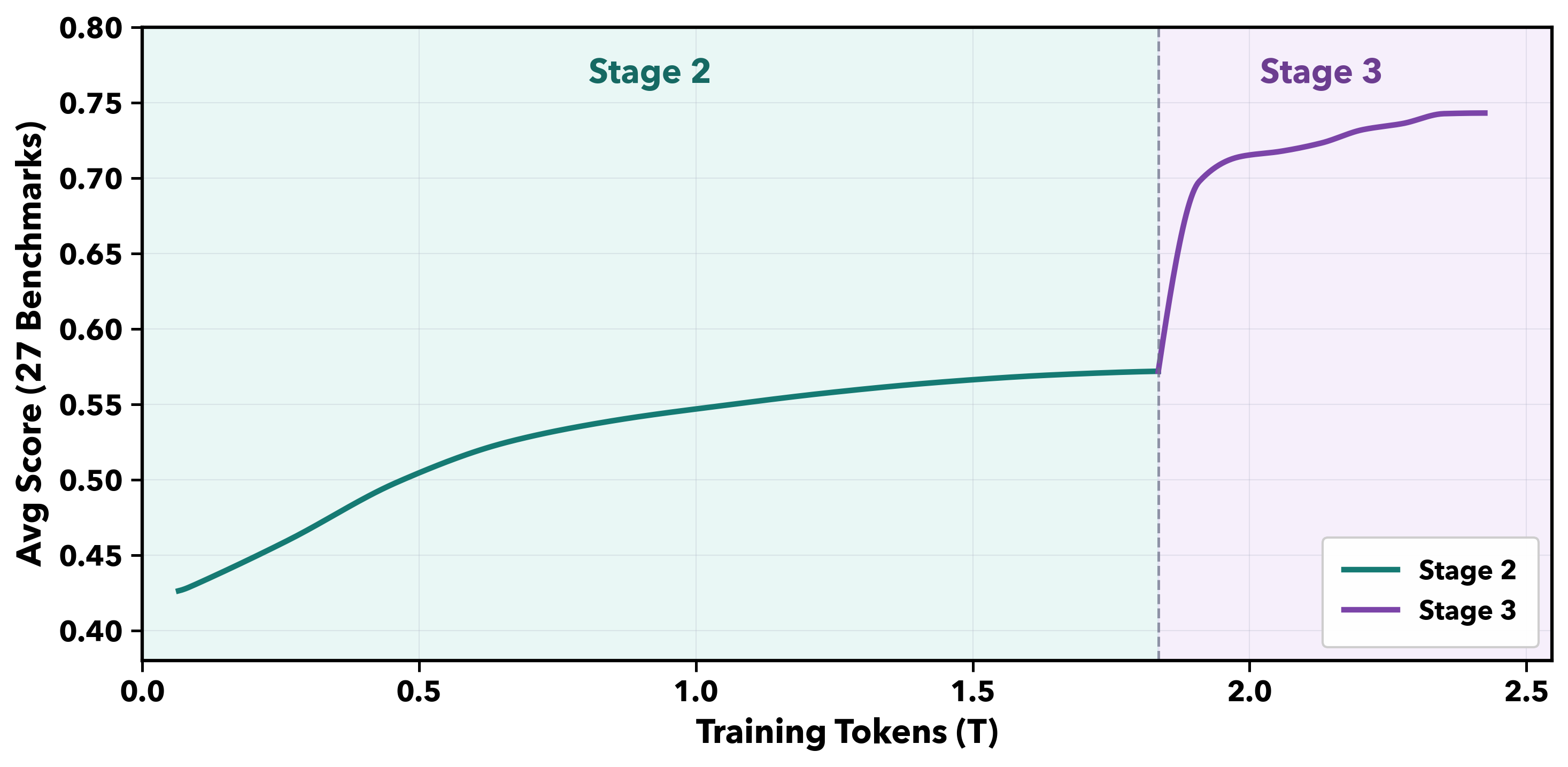}    
    \caption{\textbf{Scalability of Pre-training.}}
    \label{fig:scaling}
  \end{subfigure}
  \hfill   
  \begin{subfigure}[b]{0.57\linewidth}
    \centering
    \includegraphics[width=\linewidth]{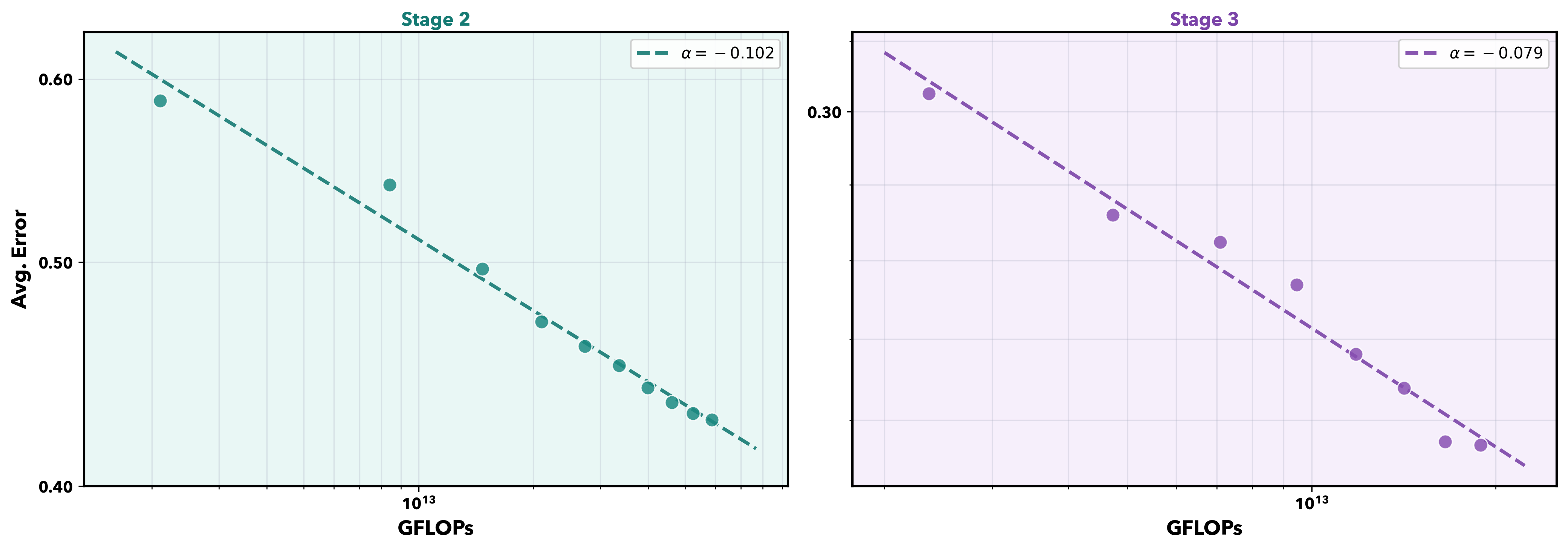}
    \caption{\textbf{Power-Law Scaling Dynamics.}}
    \label{fig:scaling_law}
  \end{subfigure}

  \caption{\textbf{Scaling analysis of VIVAS.} (a) The evolution of average performance across 27 multimodal benchmarks as a function of training tokens. The results reveal steady and consistent capability improvements during training across different data regimes, indicating that the model effectively exploits large-scale data without showing signs of empirical saturation.
(b) Examination of the mean evaluation error ($1 - \text{Score}$). The optimization dynamics in both Stage 2 and Stage 3 follow neural scaling laws, with scaling exponents estimated at $\alpha \approx 0.102$ and $\alpha \approx 0.079$, respectively.}  
  \label{fig:combined_scaling_analysis}
\end{figure*}

We demonstrate the scalability and stability of our 2.4T-token pre-training regime through empirical scaling laws. Figure~\ref{fig:scaling} shows a monotonic increase in multimodal performance (averaged over 27 benchmarks) from 0.43 to >0.74, spanning both Stage 2 (Foundation) and Stage 3 (Task Adaptation) without early saturation. 

By mapping evaluation error against compute (GFLOPs), Figure~\ref{fig:combined_scaling_analysis} confirms a predictable power-law decay, $L(C) \propto C^{-\alpha}$. The model exhibits a steep scaling exponent of $\alpha \approx 0.102$ during Stage 2, indicating rapid information absorption from mixed corpora. In Stage 3, the exponent naturally moderates to $\alpha \approx 0.079$. This sustained log-linear scaling underscores the sample efficiency of our framework, effectively converting massive computational investment into predictable capability gains.

\section{Conlcusion}

In this paper, we study the limitation of existing Vision–Language Models (VLMs) in fine-grained visual perception and attribute it to the text-dominant optimization bias in standard training paradigms. To address this issue, we introduce visual supervision into the pretraining stage and show that the unified token space paradigm enables stable and effective training for vision–language autoregressive models. Based on this paradigm, we propose a dense-structural-semantic vision tokenizer that integrates dense structural features from DINOv3 and semantic representations from SigLIP2 to construct a modality-aligned visual vocabulary. Building on this tokenizer, we develop VIVAS, a vision–language pretraining framework with unified supervision over visual and textual tokens. Extensive experiments demonstrate that VIVAS consistently improves multimodal understanding while introducing minimal inference overhead.

\section*{Acknowledgments}
This work was partly supported by the Special Foundations for the Development of Strategic Emerging Industries of Shenzhen(No.KJZD20231023094700001) and the Shenzhen-Tsinghua Special Project for Fundamental \& Frontier Research in Artificial Intelligence(No.AI2026018).

\bibliographystyle{splncs04}
\bibliography{main}
\end{document}